# A Neural Network Assembly Memory Model Based on an Optimal Binary Signal Detection Theory

Petro M. Gopych

V.N.Karazin Kharkiv National University, Kharkiv 61077 Ukraine
pmg@kharkov.com

**Abstract.** A ternary/binary data coding algorithm and conditions under which Hopfield networks implement optimal convolutional and Hamming decoding algorithms has been described. Using the coding/decoding approach (an optimal Binary Signal Detection Theory, BSDT) introduced a Neural Network Assembly Memory Model (NNAMM) is built. The model provides optimal (the best) basic memory performance and demands the use of a new memory unit architecture with two-layer Hopfield network, $N$-channel time gate, auxiliary reference memory, and two nested feedback loops. NNAMM explicitly describes the dependence on time of a memory trace retrieval, gives a possibility of metamemory simulation, generalized knowledge representation, and distinct description of conscious and unconscious mental processes. A model of smallest inseparable part or an "atom" of consciousness is also defined. The NNAMM's neurobiological backgrounds and its applications to solving some interdisciplinary problems are shortly discussed. BSDT could implement the "best neural code" used in nervous tissues of animals and humans.

## 1 Introduction

Methods previously developed for nuclear spectroscopy data processing [1], [2] became a ground for a way of strict definition and numerical computation of basic memory performance as a function of the intensity of cue [3]. Next, it was shown that coding/decoding approach [3] is optimal one and on this basis a Neural Network Assembly Memory Model (NNAMM) has been proposed [4]. In addition to optimal (the best) basic memory performance, receiver operating characteristic (ROC) curves, and etc [3], [4], [5], the NNAMM explicitly describes the dependence on time of a memory trace retrieval, gives a possibility of one-trial learning, memory and metamemory simulation, generalized knowledge representation, and distinct description of conscious and unconscious mental processes. It has been shown that the model demands a new memory unit architecture with regular *Hopfield neural network*, auxiliary *reference memory*, and *two nested feedback loops* and that such a unit may be viewed as a model of a smallest inseparable part or an "atom" of consciousness. NNAMM is strongly supported by many neurobiological arguments some of them were introduced as counterparts to neural network (NN) memory elements for the first time. Among many other more obvious applications, the model provides a possibility of building quantitative NN models of some mental phenomena which were up to present outside of the recent computational theories (for example, the tip-of-the-tongue phenomenon [6] or feelings and emotions [7]).

## 2 An Optimal Binary Signal Detection Theory

An optimal Binary Signal Detection Theory (BSDT) sketched below is based on a data coding/decoding approach [3] where initial data are represented by ternary vectors with their components' possible magnitudes from the triple set –1, 0, 1. Each ternary vector, or pattern of signals, simulates specific pattern of simultaneously (within a given time window) acting nerve impulses, action potentials, or "spikes" and its every component simulates the fact of the absence (0) or presence (±1) of spike affecting on excitatory (+1) or inhibitory (–1) synapse of the target neuron (i.e., the sign of the vector's nonzero component indicates the specific spike's further assignment). The number of the ternary vector's components is large, because central nervous system contains very much neurons, but most of them should be equal to 0, because most neurons are "silent" or "dormant" at the moment. This situation respects to the data *sparse* coding and we refer to initial ternary vectors' space dimension as $N_{sps}$. Next, it is assumed that silent neurons do not carry any information of current particular interest and, therefore, they should be excluded from the further current particular consideration. Such a process (under which ternary vector's zero components are excluded) is virtually a transition from sparse- to *dense*-coding representation. We suppose that ternary vectors are transformed into binary ones at the stage of initial data *preprocessing* when only those spikes from the initial data flood are selected which are fallen into a time-coincidence window of particular cell assembly allocated by a dynamic spatiotemporal synchrony mechanism (see Sect. 4). The dimension $N_{dns}$ of densely coded binary *feature* vectors obtained as a result is much less then $N_{sps}$, $N_{dns} \ll N_{sps}$. In fact, each feature vector is *quasi-binary* one because its spinlike (±1) components cannot be shifted to other (0, 1) binary representation by the redefinition of coupling constants and thresholds and it can manifest but in spite of that does not manifest its zero components (although they are important for memory impairment definition). Below only $N$-dimensional ($N = N_{dns}$) quasi-binary vectors are considered but, for short, the preposition "quasi" will be omitted [4].

A method for spinlike (±1) data production was proposed in ref. 2 where line radiation spectra (a kind of half-tone images) were considered. Moreover, it was found that in the course of initial gradual data binarization (and their compression simultaneously), there is no loss of information important for the binarized data following processing. Hence, there exists a broad class of problems (local feature discrimination in noise) where spinlike data described naturally occur. Also we found [4] that population bursts [8] and/or distributed bursts [9] of spiking activity discovered in animal nervous tissues may be viewed as counterparts to spinlike vectors needed for BSDT.

### 2.1 Data Coding

We denote a vector with components $x^i$, $i = 1,...,N$, which magnitudes may be –1 or +1, as *x*. It can carry $N$ bits of information and its dimension $N$ is *the size* of a local receptive field of the NN (and convolutional) feature discrimination algorithm [2] or *the size* of a future NN memory unit (Sect. 2.2). If $x$ represents information stored or that should be stored in an NN then we term it *reference vector* $x_0$. If the signs of all components of $x$ are randomly chosen with uniform probability then that is *random vector* $x_r$ or *binary "noise"*. We define also a *damaged reference vector* $x(d)$ with *damage degree* of $x_0$ $d$ and components

$$x_i(d) = \begin{cases} x_0^i, \text{if } u_i = 0, \\ x_r^i, \text{if } u_i = 1 \end{cases} \quad i = 1,...,N \quad (1)$$

where $u_i$ denotes marks whose magnitudes, 0 or 1, are randomly chosen with uniform probability and fixed $d$:

$$d = \sum u_i / N, \quad i = 1,...,N. \quad (2)$$

It is clear that $x_r$ takes priority over $x_0$. If the number of marks $u_i = 1$ is $m$ then $d = m/N$; $0 \leq d \leq 1$, $x(0) = x_0$, and $x(1) = x_r$.

If $x(d)$ contains a part $q$ of undamaged information about $x_0$ then $x(d) = x(1 - q)$ where $q$ is *intensity of cue* or *cue index* [3]:

$$q = 1 - d. \quad (3)$$

It is clear that $q = 1 - m/N$, $0 \leq q \leq 1$, and $x(q) = x(1 - d)$; $d$ and $q$ are discrete values.

If $d = m/N$ then the number of different vectors $x(d)$ is $2^m C^N_m$, $C^N_m = N!/(N - m)!/m!$; if $0 \leq d \leq 1$ then complete *finite* set of vectors $x(d)$ consists of $\sum 2^m C^N_m = 3^N$ elements ($m = 0,1,...,N$). For specific $x(d)$ $d$ and $q$ may be interpreted as *noise-to-signal ratio* and *signal-to-noise ratio*, respectively. Cue index $q$ is also a degree of similarity or *correlation coefficient* between $x_0$ and $x(d)$. In spite of the fact that in Eq. 1 vectors $x_0$ and $x_r$ are combined unequally, for each $x(d)$ signal and noise are *additive* in that sense that $q + d = 1$. Vectors $x(d)$ do not contain zero components although they may contain a fraction $d$ of noise which is their natural and inherent part [4].

## 2.2 Data Decoding

Now for the data coding introduced we define decoding rules, an algorithm for extracting $x_0$ from $x_{in} = x(d)$ interpreted as a sample of pure noise or as $x_0$ distorted by noise with noise-to-signal ratio $d$. For this purpose we consider a two-layer autoassociative NN with $N$ McCalloch-Pitts model neurons in its entrance (and exit) layer and suppose that all cells from the NN entrance layer are linked to all exit layer cells according to the "all-to-all" rule.

Following ref. 10 for learned NN, we define elements $w_{ij}$ of *synapse matrix* $w$ as

$$w_{ij} = \eta \, x_0^i x_0^j \quad (4)$$

where $\eta > 0$ is a *learning parameter* (below $\eta = 1$); $x^i_0$, $x^j_0$ are the $i$th and the $j$th components of $x_0$, respectively. Hence, using the information (vector $x_0$) that should be stored in the NN, Eq. 4 defines $w$ unambiguously. We refer to $w$ as a perfectly learned NN and stress the crucial importance of the fact that it remembers *only one* pattern $x_0$

(the available possibility of storing other memories in the same NN was *intentionally disregarded*). It is also assumed that input vector $x_{in}$ is decoded (reference vector $x_0$ is extracted) successfully if learned NN transforms an $x_{in}$ into the output vector $x_{out} = x_0$. The transformation algorithm is the following.

For the $j$th NN exit layer neuron, an *input signal* $h_j$ is defined as

$$h_j = \sum w_{ij} v_i + s_j \tag{5}$$

where $v_i$ is an *output signal* of the $i$th neuron of the NN entrance layer; $s_j = 0$.

For the $j$th NN exit layer neuron, output signal $v_j$ (the $j$th component of $x_{out}$) is calculated according to rectangular response function (signum function or 1 bit quantifier) with the model neuron's triggering threshold $\theta$:

$$v_j = \begin{cases} +1, & if \quad h_j > \theta \\ -1, & if \quad h_j \leq \theta \end{cases} \tag{6}$$

where for $h_j = \theta$ the value $v_j = -1$ was *arbitrary* assigned.

If $h_i = x_{in}^i$ then $v_i = x_{in}^i$. Of this fact and Eqs. 4 and 5 for the $j$th exit layer neuron we have: $h_j = \sum w_{ij} x_{in}^i = \eta x_0^j \sum x_0^i x_{in}^i = \eta x_0^j Q$ where $Q = \sum x_0^i x_{in}^i$ is a convolution of $x_0$ and $x_{in}$. The substitution of $h_j = \eta x_0^j Q$ into Eq. 6 gives that $x_{out} = x_0$ and an input vector $x_{in}$ is decoded (reference vector $x_0$ is extracted) successfully if $Q > \theta$ (if $\eta \neq 1$ then $Q > \theta/\eta$). Since for each $x_{in}$ exists such a vector $x(d)$ that $x_{in} = x(d)$, inequality $Q > \theta$ can also be written as a function of $d = m/N$:

$$Q(d) = \sum_{i=1}^{N} x_0^i x_i(d) > \theta \tag{7}$$

where $\theta$ is the threshold value of $Q$ or the model neuron's triggering threshold. Hence, above NN and convolutional decoding algorithms are equivalent [2] although Ineq. 7 is valid only for perfectly learned intact NNs.

For example, directly we can find that $Q = N - 2D$ and $D = (N - Q)/2$. Here $D$ is Hamming distance between $x_0$ and specific $x(d)$ or the number of their corresponding bits which are different. Thus, along with Ineq. 7 the inequality $D < (N - \theta)/2$ is also valid. Moreover, $Q(d)$ can merely be interpreted as an expression for convenient computation of $D(d)$. That means that Hamming (convolutional) decoding algorithm or Hamming linear classifier directly discriminates the patterns $x_{in} = x(d)$ which are more close to $x_0$ than Hamming distance given. Hence, NN, convolutional, and Hamming distance decoding algorithms mentioned are equivalent [4], [5].

As Hamming decoding algorithm is the best (optimal) in the sense of statistical pattern recognition quality (that is no other algorithm cannot outperform it) [11], NN and convolutional algorithms described above are also optimal (the best) in that sense. Moreover, similar decoding algorithms based on locally damaged NNs may also be optimal, at least if their damages are not catastrophically large [4].

### 2.3 Data Decoding Quality Performance

The optimality of the BSDT was just demonstrated and now we will describe quantitatively its decoding algorithm quality performance. For such an algorithm, they are

$P_\theta(d)$, the probability of correct decoding conditioned under the presence or absence of $x_0$ in the data analyzed vs. damage degree $d$ or intensity of cue $q = 1 - d$, and/or $P_d(F)$, the probability of correct decoding vs. model neuron triggering threshold $\theta$ or false alarm probability $F$ (that is receiver operating characteristics or ROC curves [12]). For unconditional (*a posteriori*) and overall probabilities of correct and false decoding derived within the framework of the BSDT see ref. 5.

**Functions $P_\theta(d)$**
The finiteness of the complete set of vectors $x(d)$ (see Sect. 2.1) makes possible to find probabilities $P_\theta(d)$ using multiple computations; convolutional (Hamming) version of the BSDT (see Sect. 2.2) allows to derive formulae for $P_\theta(d)$ analytically.

*Calculation of $P_\theta(d)$ by multiple computations*
Probabilities $P_\theta(d)$ are calculated as $P_\theta(d) = n_\theta(d)/n(d)$ where $n(d)$ is a number of different randomly generated inputs $x_{in} = x(d)$ with constant damage degree $d$; $n_\theta(d)$ is a number of inputs which lead to the emergence of the NN's response $x_{out} = x_0$. For small $N$ $P_\theta(d)$ can be calculated *exactly* (Figs. 1a and 3) because $n(d)$ is small and all possible inputs may be taken into account: $n(d) = 2^m C^N_m$ and $d = m/N$ ($m \leq N$ is a number of marks $u_i = 1$ in Eq. 2). For large $N$ $P_\theta(d)$ can be estimated by multiple computations approximately but *with any given accuracy* [3], [4].

*Analytical formulae for $P_\theta(d)$*
A formula for calculating the probability of correct decoding vectors $x_{in} = x(d)$ by intact NN perfectly learned to discover pattern $x_0$ was found in ref. 5 as

$$P(m, N, \theta) = \sum_{k=0}^{k\max} C^m_k / 2^m \qquad (8)$$

where $C^m_k$ denotes binomial coefficient; if $k\max \leq m$ then $k\max = m$ else $k\max = k\max_0$ and $k\max_0$ is defined as

$$k\max_0 = \begin{cases} (N - \theta - 1)/2, & \text{if } N \text{ is odd} \\ (N - \theta)/2 - 1, & \text{if } N \text{ is even}. \end{cases} \qquad (9)$$

Since $0 \leq k\max \leq m \leq N$, if $N$ is odd then $-(N+1) \leq \theta \leq N-1$ and if $N$ is even then $-(N+2) \leq \theta \leq N-2$. As one can see from Fig. 1, $P(m,N,\theta) = P_\theta(m/N) = P_\theta(d) = P_\theta(1-q)$; $P_\theta(d) = P_F(d)$.

For each decoding algorithm based on a perfectly learned NN with specific distribution of its local damages, the expression for $P_\theta(d)$ should be derived separately.

**ROC Curves**
ROC or $P_d(F)$ curves (Fig. 1b) are merely the other form of plotting functions $P_\theta(d) = P_F(d)$ (Eqs. 8, 9 and Fig. 1a). Hence, complete families of curves $P_d(F)$ and $P_F(d)$ carry *the same* information about the BSDT's decoding algorithm quality performance while pairs of particular curves $P_d(F)$ and $P_F(d)$ contain *different* information (each specific pair of these curves has a single common point). We see that ROCs depend on the argument $F$ and *only one* parameter, $d = m/N$.

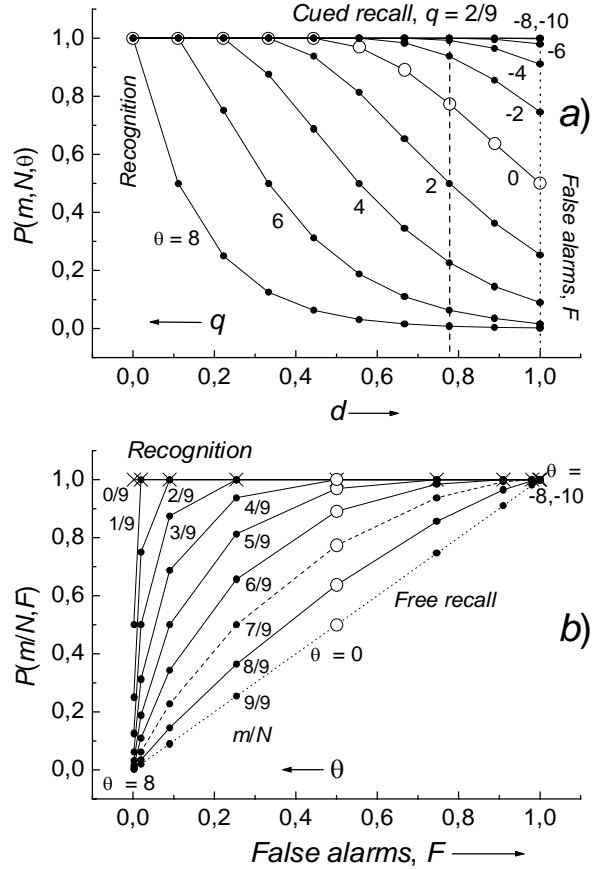

**Fig. 1.** a) $P(m,N,\theta) = P_\theta(d)$ vs. $d = m/N$ (or $q = 1 - m/N$) and $\theta$. Right-most point of each curve for each model neuron's triggering threshold $\theta$ represents respective value of the false alarm probability $F = P_\theta(1)$. b) $P(m/N,F) = P_d(F)$ vs. $F$ and $m/N$. Points related to the same value of $m/N$ (the same initial data quality) are connected with straight lines and make specific ROC curve. In both panels values of $P$ for $q = 0$ (*dotted lines*), $q = 2/9$ (*dashed lines*), $\theta = 0$ (*open circles*), and $m = 0$ (*crosses*) are picked out. Examples for ideally learned NN with $N = 9$.

## 3 A Neural Network Assembly Memory Model

The basic idea of the present work is to build a NN memory model from simple objects with simple known properties defined within the optimal BSDT introduced.

In such a way a *Neural Network Assembly Memory Model* (NNAMM) or *assembly memory* for short was built [4] from similar interconnected (associated) and equal in right *Assembly Memory Units* (AMU) each of which respects to particular feature of a stimulus and is represented by a particular cell assembly allocated in the brain by means of a dynamic spatiotemporal synchrony mechanism (Sect. 4). Each AMU con sists of the traditional *Hopfield NN* and some functionally new and arranged in a new manner elements among which there are *N-channel time gate, reference memory*, and *two nested feedback loops*. It is assumed again that specific NN remembers *only one* memory trace $x_0$ which is retrieved successfully if an input $x_{in} = x(d)$ initiates the emergence of the output $x_{out} = x_0$. In contrast to data decoding *one-step* process, NNAMM implies a *many-step* memory retrieval. Due to the optimality of the initial decoding algorithm, *memory retrieval process* and its *performance* are also optimal.

Neurobiological plausibility of the NNAMM in a whole and its separate building blocks are discussed in Sect. 4. The comparison of the NNAMM with some other memory approaches (Hopfield NNs, sparsely coded Kanerva model, convolutional and modular/structured/compositional memories) can be found in [4].

### 3.1 An Assembly Memory Unit

An AMU (Fig. 2) consists of *blocks 1-6* and their internal and external *pathways* and *connections* designed for propagation of synchronized groups of signals (vectors $x(d)$) and asynchronous control information, respectively.

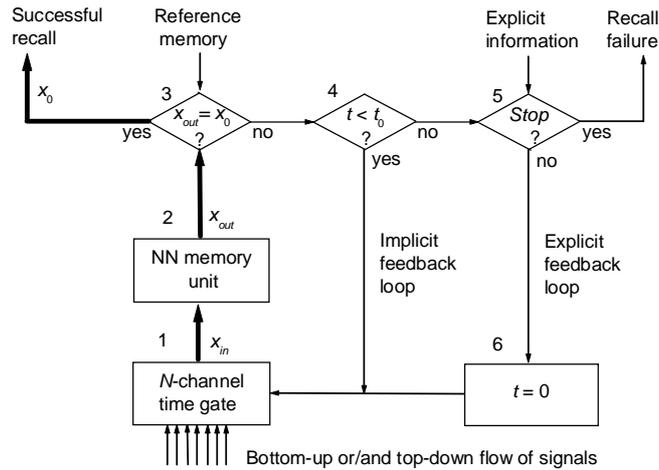

**Fig. 2.** The flow chart of a particular assembly memory unit and its short-distance environment. Pathways and connections are shown in thick and thin arrows, respectively.

*Block 1* (a kind of *N*-channel time gate) transforms initial ternary sparsely coded vectors into binary (spinlike) densely coded ones. Here from the flood of generally asynchronous input spikes, synchronized pattern of signals in the form of *N*-dimensional *feature vector* $x_{in}$ is prepared. *Block 2* is a NN memory unit learned according to Eq. 4 where each input $x_{in}$ is transformed into an output $x_{out}$ according to Eqs. 5 and 6. *Block 3* performs the comparison of vector $x_{out}$ just now emerged with *reference vector* (*trace*) $x_0$ from *reference memory* (see below). If $x_{out} = x_0$ then the retrieval is successful and it is finished. In opposite case, if current *time of retrieval t* is less then its maximal value $t_0$ (this fact is checked in *block 4*) then the internal or

*implicit feedback loop* 1-2-3-4-1 is activated (see below), retrieval starts again from *block 1*, and so forth. If $t_0$, a parameter of time dependent neurons (Sect. 4), was found as insufficient to retrieve $x_0$ then *block 5* examines whether an external reason exists to continue retrieval. If it is then the external or *explicit feedback loop* 1-2-3-4-5-6-1 is activated (see below), the count of time begins anew (*block 6*), and internal cycle 1-2-3-4-1 is repeated again with the given frequency $f$, or time period $1/f$, while $t < t_0$.

An AMU may be viewed as a kind of finite state "neural microcircuit" or "chip" for "anytime" or "real-time" computing [13] but in contrast to [13] computations are performed in binary form and the neural code used is explicitly known (Sect. 2).

**Reference Memory**
A memory trace $x_0$ is stored simultaneously in a particular NN memory (*block 2*) and in its auxiliary *reference memory* (RM) here introduced (Fig. 2). RM may be interpreted as a *tag* or a *card* of the memory record in a memory catalog and performs two interconnected functions: *verification* of current memory retrieval results (*block 3*), and *validation* of the fact that a particular memory actually exists in the long-term memory store (at the stage of memory activation). Thus, specific RM is a part of "memory about memory" or "*metamemory*." In other words, "memory about memory" means "knowledge about memory." Hence, RM is also a kind of generalized (in the form of *N*-bit binary code) *knowledge representation*. In contrast to regular NN memory which is a kind of *computer register* and is conventionally associated to real biological network, particular RM is a kind of *slot* devoted to the *comparison* of a current vector $x_{out}$ with the reference pattern $x_0$ and may be associated to a specific integrate-and-fire neuron, e.g. [13]. RM provides also a possibility of synergistic coding/decoding: thanks to RM information extracted from a two-layer autoassociative NN with $N$ neurons in a layer may be greater then that carried by $N$ independent individual neurons. For details and description of the other AMU elements see [4].

**Two Nested Feedback Loops**
All elements of the internal loop 1-2-3-4-1 (Fig. 2) run routinely in an automatic regime and for this reason they may be interpreted as respected to *implicit* (unconscious) memory. That means that under NNAMM all neural operations at synaptic and NN memory levels are unconscious. External loop 1-2-3-4-5-6-1 is activated in an unpredictable manner because it relies on external (environmental and, consequently, unpredictable) information and in this way provides unlimited diversity of possible memory retrieval modes. For this reason an AMU can be viewed as a particular *explicit* (conscious) memory unit. An external information in *block 5* used can be thought of as an *explicit* or conscious one (for distinctions between implicit and explicit memories see [14]); *error detector neurons* (Sect. 4) may participate in *block 5* construction and may be interpreted as related to *neural correlates of consciousness* [15]. Hence, only at the level of a particular AMU, a possibility arises to take *explicit* (conscious) factors into account and, consequently, particular AMU is a smallest inseparable part or an "atom" of all possible explicit memories. Thus, an AMU may be used as a building block for construction of all the high-level conscious memories and conscious brain functions in general. These suggestions are consistent with the notion on the modularity of consciousness or the multiple small-scaled consciousness [16].

## 3.2 Numerical Example

NN local damages and respective memory impairments (Fig. 3) may be caused by natural reasons (that is a natural *forgetting*) or by the brain's *trauma* or *decease* [4].

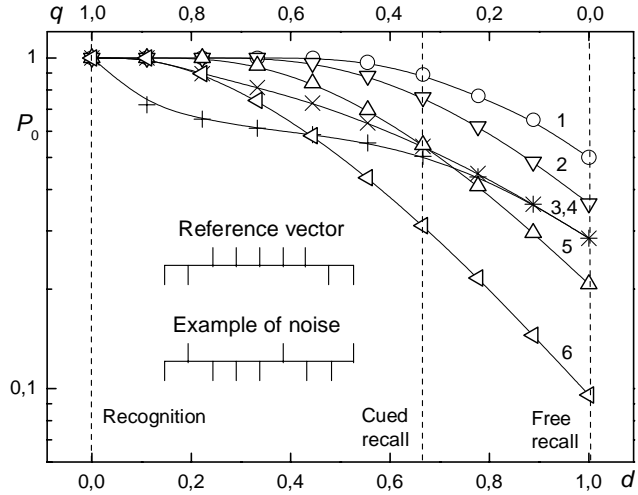

**Fig. 3.** Probabilities $P_0(d) = P_0(1 - q)$ for undamaged (*curve 1*) and damaged (*curves 2-6*) perfectly learned NNs with $N = 9$ and $\theta = 0$. Exact calculation results (*different signs*) are connected with interpolation curves. For curves 3 and 4 the number of "killed" entrance neurons is $N_k = 4$; for curves 2, 5, and 6 the number of links disrupted between entrance and exit neurons is $N_d = 10$. Dashed lines denote free recall ($q = 0$), recognition ($q = 1$), and cued recall ($q = 1/3$) probabilities (for fixed $d$ the less $P_0(d)$ the greater memory impairment is). In the insertion components $\pm 1$ of the vector $x_0$ and a sample of noise are shown as hyphens located above (+1) or below (–1) horizontal lines. In Figs. 1 and 3 open circles respect to the same values of $P_0(d)$.

## 3.3. Basic Memory Performance

Since we define probabilities of the identification of $x_0$ in $x_{in} = x(d)$ as $P_\theta(d)$, $P_\theta(d)$ is also the probability of retrieving $x_0$ from an NN tested by a series of $x_{in}$ with fixed cue $q$. The vector $x_{in} = x_0$ is recognized as $x_0$ with the probability $P_\theta(0) = 1$; noise $x_{in} = x_r$ is recognized as $x_0$ randomly with the probability $P_\theta(1) = \alpha < 1$. Hence, $\alpha \leq P_\theta(d) \leq 1$.

Below we consider the recall/recognition algorithm introduced as a *test* for the distinction of simple alternative statistical hypotheses.

Let *null hypothesis* $H_0$ be such that a statistical sample $x_{in} = x(d)$ is a sample of noise, i.e. $d = 1$ and $x_{in} = x_r$. By definition, the probability $\alpha = P_\theta(1)$ of rejecting $H_0$ is *test significance level, Type-1 error rate,* and *conditional probability of false discovery*. Noise does not contain any information about $x_0$ and, therefore, $\alpha$ is also the probability of the recall $x_0$ without any cue or *free recall probability*. Thus, $\alpha$ is simultaneously *test significance level, Type-1 error rate, conditional probability of false discovery* (*false alarm probability*), and *free recall probability*.

*Alternative hypothesis* $H_1$ is such that statistical sample $x_{in}$ is $x_0$ with damage degree $d$, i.e. $x_{in} = x(d)$, $0 \leq d < 1$. If the probability of rejecting $H_1$, where it is true, is $\beta$ then $\beta$ is *Type-2 error rate*. Under the same condition probability of taking $H_1$ is *test power* $M = 1 - \beta$ and, since for a sample $x(d)$ NN response $x_0$ emerges with probability $P_\theta(d)$, $M = P_\theta(d)$. Each statistical sample $x(d)$ contains a part $q$ of information about $x_0$ which "reminds" the learned NN about $x_0$ and in this way "helps" it to recall $x_0$. Therefore, $P_\theta(d)$ is also the probability of the recall $x_0$ with a cue $q = 1 - d$ or *cued recall probability*. Thus, $P_\theta(d)$, $0 < d < 1$, is simultaneously *test power*, *conditional probability of true discovery*, and *cued recall probability*. $P_\theta(0)$, $d = 0$, is *recognition probability*.

## 4 The NNAMM's Neurobiological Background

For the NNAMM substantiation besides Hebbian synaptic rule, brain's feedback loops, bottom-up and top-down pathways, we use some nontraditional neurobiological arguments. Because of space limitations, we refer to [4] for details and references.

Lately, convincing experimental evidences have been obtained that when a subject performs an attention related cognitive task, cortical neurons within a small group synchronize their activity with the precision of about 10 ms in gamma-band frequency range, ~ 40 Hz, — that is *dynamic spatiotemporal synchronization* phenomenon. Within the NNAMM, a synchrony mechanism allocates particular cell assembly representing an assembly memory and chooses its respective pattern of signals, a message addressed to particular AMU in the form of the $N$-dimensional vector $x(d)$ [4].

*The first precisely aligned spike pattern* of neuronal responses from synchronous cell assembly (early precise firing phenomenon), *population bursts* [8], and *distributed bursts* [9] of spiking neurons constitute a strong neurobiological ground for vectors $x(d)$. This assertion is consistent with Shannon information theory result that in the case of non-interactive neurons mutual information is carried entirely in firing rate, but for the correlational neuron population the information is fully conveyed by the correlational component with no information in the firing rate [17].

The size $N$ of a particular AMU is estimated as ~100. The reasons are the size of fundamental signaling unit in cerebral cortex, the number of thalamic axons which project to a given cortical neuron, the number of relay neurons in cat's visual system which is sufficient to satisfactorily reconstruction of natural-scene movies, the size of the population of cortical motor neurons which can control with good quality one- and three-dimensional movements of robot arms, and the size of the pool of spiking neurons stable propagating in complex cortical networks. The later finding supports also our assumption that stable propagation of vectors $x(d)$ along external and internal pathways in Fig. 2 is neurobiologically well motivated.

As a natural time scale for the memory retrieval mechanism (Sect. 3.1), *decay period* $t_0$ of *time dependent neurons* is used. Such neurons start their activity when transient stimulus occurs and afterward decrease their spiking rate according to linear law during a period $t_0$ that can vary from tens milliseconds to tens seconds, e.g. [18].

In some subcortical areas it were discovered *"error detector"* neurons [19], neuronal populations which selectively change their firing rates only when errors were made in *cognitive* tasks. In the NNAMM they are used to design *block 5* in Fig. 2.

## 5 Some Applications

The recognition algorithm based on Eqs. 5 and 6 was implemented as computer code *PsNet* [2] designed for solving the problem of local feature discriminations in one-dimensional half-tone images ("line spectra") and for full quantitative description of psychometric functions obtained [1] as a result of testing human visual system.

For damaged NNs their free and cued recall probabilities ($P_0(d)$, $0 < d \leq 1$) are less in general than for undamaged ones but for all NNs their recognition probabilities are the same, $P_0(0) = 1$ (Fig. 3). Thus, for damaged NNs, NNAMM predicts the retain of their recognition ability and the impairment of their free and cued recalls. These agree with the data on *episodic memory performance* in patient with frontal lobe local damages [20]. Since free recall, cued recall, and recognition are special cases of a single unified mechanism of memory retrieval, the model supports the assumption about a *close relationship* between recall and recognition [21] and is, in first approximation, not consistent with the prediction that they depend on different brain systems [22].

NNAMM became a ground for an NN model of the *tip-of-the-tongue* (TOT) phenomenon (in Fig. 3 curve 4, with a semi-plateau, corresponds to particular TOT state). The model makes possible to define and calculate the TOT's strength and appearance probabilities, to join memory, psycholinguistic, and metamemory TOT's analyses, to bridge speech error and naming chronometry traditions in TOT research [6].

NNAMM provides distinct description of conscious and unconscious mental processes (Sect. 3.1). Taking this fact into account and using recent emotion theories, conceptual and quantitative NN models of feelings and emotions were proposed and applied to description of the *feeling-of-knowing* [7]. The model gives a chance to explain different feeling, emotion, or mood phenomena both in animals and humans, but its main inference is that emotions do not distinct conscious and unconscious mental processes but only create respective emotional background for them [7].

Empirical ROCs obtained in item and associative recognition tests [23] were quantitatively described and the values of the intensity of cue for some specific experiments were estimated [5]. It was also shown [5] that ROCs might be excluded from the list of findings underpinning dual-process models of recognition memory.

## 6 Conclusions

An optimal BSDT, based on it NNAMM, some their features, and applications have been discussed. For assembly memory activation, one-trial memory learning, relations between active and passive memory traces, free NNAMM parameters, memory impairments, comparison to other memory models, additional references, and discussions see [4]. All the NNAMM's advantages are caused by advantages of the BSDT. We even hypothesize that BSDT could implement "the best neural code" [9]. The reasons are four-fold [4]: NNAMM, information theory, neurophysiology experiments, and computer modeling of NNs with nearly real neurophysiology parameters.

I am grateful to Health Internetwork Access to Research Initiative (HINARI) for free on-line access to recent full-text journals, anonymous referee for comments to the initial manuscript, and my family and my friends for their help and support.